%% file: main.tex
\documentclass{article} 
\usepackage{iclr2021_conference,times}

\input{math_commands.tex}

\usepackage{dsfont}
\usepackage{hyperref}
\usepackage{url}
\usepackage{graphicx}
\usepackage{float}
\usepackage[noline,boxed]{algorithm2e}
\title{Fractional Transfer Learning for \\ 
Deep Model-Based Reinforcement Learning}

\author{Remo Sasso * \\
Dept. Artificial Intelligence\\
University of Groningen\\
\texttt{\href{mailto:r.sasso@student.rug.nl}{r.sasso@student.rug.nl}}\\
\And
Matthia Sabatelli \\
Dept. Artificial Intelligence\\
University of Groningen\\
\texttt{\href{mailto:m.sabatelli@rug.nl}{m.sabatelli@rug.nl}}\\
\And
Marco A. Wiering \\
Dept. Artificial Intelligence\\
University of Groningen\\
\texttt{\href{mailto:m.a.wiering@rug.nl}{m.a.wiering@rug.nl}}\\
}
\iclrfinalcopy 
\begin{document}

\maketitle

\begin{abstract}
Reinforcement learning (RL) is well known for requiring large amounts of data in order for RL agents to learn to perform complex tasks. Recent progress in model-based RL allows agents to be much more data-efficient, as it enables them to learn behaviors of visual environments in imagination by leveraging an internal World Model of the environment. Improved sample efficiency can also be achieved by reusing knowledge from previously learned tasks, but transfer learning is still a challenging topic in RL. Parameter-based transfer learning is generally done using an all-or-nothing approach, where the network's parameters are either fully transferred or randomly initialized. In this work we present a simple alternative approach: fractional transfer learning. The idea is to transfer fractions of knowledge, opposed to discarding potentially useful knowledge as is commonly done with random initialization. Using the World Model-based Dreamer algorithm, we identify which type of components this approach is applicable to, and perform experiments in a new multi-source transfer learning setting. The results show that fractional transfer learning often leads to substantially improved performance and faster learning compared to learning from scratch and random initialization.
\end{abstract}

\section{Introduction}
Reinforcement learning (RL) \citep{Wiering_undated-hb,Sutton2018-yv} is a branch of artificial intelligence research that focuses on developing algorithms which enable agents to learn to perform tasks in an environment they are situated in. RL agents learn to improve their behavior by means of maximizing the rewards they receive from an environment. Rewards result from transitions in the environment that are initiated by the agent taking actions. This type of learning can be a powerful tool for creating intelligent agents that become able to master complicated tasks, occasionally resulting in superhuman capabilities. For instance, the RL based AlphaZero algorithm learned to master the games of Go, shogi, and chess at a superhuman level \citep{Schrittwieser2019-df}. However, in order for RL agents to obtain these capabilities, the required amount of interaction with the environment can be enormous. An important issue in RL research is therefore sample efficiency, defined as how much interaction is necessary for the agent to master the task at hand. 

In model-based RL (MBRL), the agents have access to an internal model of the world \citep{DBLP:journals/corr/abs-2006-16712}. This model allows the agent to predict transitions in the environment that result from its actions, without having to interact with the environment. The model therefore allows agents to perform planning, as well as generating internal trajectories that can be used for learning, resulting in improved sample efficiency. However, the effectiveness of this type of approach depends on the quality of the model, which often needs to be learned. Better sample efficiency can also be accomplished by reusing knowledge of previously learned tasks. Transfer learning (TL) can therefore be a key component in advancing the field of RL research, but as of writing TL is still an emerging topic in RL \citep{Zhu2020-jj}.

Recently, MBRL has significantly progressed by making use of Deep Learning (DL) techniques, resulting in state-of-the-art algorithms for both discrete and continuous benchmarks \citep{Schrittwieser2019-df, Hafner2019-ab, Hafner2020-ry}. An important advancement in terms of sample efficiency was that of World Models, which allow MBRL agents to learn behaviours by imagining trajectories of environments with high-dimensional visual observations \citep{Ha2018-jd}. This type of approach provides a strong foundation for TL and multi-task learning, as World Models are of generalizable nature for different environments sharing the same physical laws \citep{Hafner2018-pq, Zhang2018-eo}. However, to the best of our knowledge, there is very few research that has been carried out for combining MBRL algorithms that make use of World Models with TL and multi-task learning.

What is commonly done in parameter-based TL, is random initialization of an output layer of a neural network. This is done in order for the pre-trained model to be able to learn a new task, as otherwise the model needs to unlearn the fit parameters of the previous task. In this work we present an alternative approach that allows fractions of parameters to be transferred instead. The reasoning behind this approach is that we don't discard all the previous knowledge contained in the output layer, but instead retain a portion of potentially useful knowledge for learning a new task. More specifically, we investigate the effectiveness of this approach in a multi-source transfer learning setting, by combining multi-task learning and TL with the state-of-the-art World Model based MBRL algorithm for continuous control tasks, Dreamer \citep{Hafner2019-ab}.

The key contributions of this work can be summarized as:
\begin{itemize}
    \item \textbf{Simultaneous multi-task transfer learning}\hspace{0.4cm}We show that simultaneously training a single World Model-based RL agent on multiple distinct visual continuous control tasks provides for a multi-source TL approach that often results in positive transfers. 
    \item \textbf{Fractional transfer learning}\hspace{0.4cm} We present a novel TL approach that allows fractions of parameters to be transferred, opposed to discarding all knowledge as is commonly done by random initialization of an output layer. This method results in substantial performance improvements and faster learning compared to learning from scratch and random initialization.
\end{itemize}

\section{Theoretical Background}
In this section we detail on theoretical concepts related to this study and discuss relevant related works. First, we provide a brief introduction to RL (Section \ref{RL}). Next, we introduce and discuss relevant literature of World Models (Section \ref{worldmodels}). Finally, we discuss how RL and World Models can benefit from TL (Section \ref{transfer}).

\subsection{Reinforcement Learning} \label{RL}
We can formalize an RL problem as a Markov Decision Process (MDP) \citep{Bellman1957-fo}, which consists of four main elements:
\begin{itemize}
    \item a set of states $S$
    \item a set of actions $A$
    \item a transition function $P(s,a,s')$ = Pr\:($s_{t+1} = s'|s_t = s,$ $ a_t = a$)
    \item a reward function $R(s,a,s')$
\end{itemize}
Here $P(s,a,s')$ denotes the probability that taking action $a \in A$  in state $s \in S$ at time step $t$ results in a transition to state $s' \in S$. When a state transition takes place, the reward function $R(s,a,s')$ yields an immediate reward corresponding to this transition. The objective in an MDP is to find an optimal policy $\pi^*(s_t) = a_t$ that maximizes the expected cumulative reward. The expected cumulative reward is defined as:
\begin{equation} \label{maxcumrew}
    {\displaystyle \mathds{E}\left[\sum _{t=0}^{T}{\gamma ^{t}R(s_{t},a_{t},s_{t+1})}\right]}
\end{equation}
Here $\gamma$ represents the discount factor (where $\gamma \in$ [0,1)), and $T$ is a random variable representing the number of time steps to account for in computing the expected cumulative reward. 

The number of states $|S|$ and actions $|A|$ may be very large or infinite, in which case neural networks have typically been used as function approximators for the past decades \citep{Tesauro1995-pe}. In more recent years, DL architectures such as Convolutional Neural Networks (CNNs) \citep{lecun2015deep} and Recurrent Neural Networks (RNNs) \citep{rnnref} also sparked the interest of the RL community, which resulted in Deep RL (DRL) research. DRL became particularly popular since the introduction of Deep Q-Networks (DQN) \citep{Mnih2015-dx}, where CNNs were used to process high-dimensional raw image data. This resulted in agents achieving a level comparable to professional human players in 49 Atari games, solely based on a reward signal and image observations. Significant advancements that followed included the iterations of AlphaGo \citep{Silver2016-df}, AlphaZero \citep{Silver2017-jq}, and eventually MuZero \citep{Schrittwieser2019-df}. The latter accomplished state-of-the-art performances in Go, Chess, Shogi, and the Atari benchmarks by making use of MBRL. DRL has also been shown to be able to master highly complex modern games, such as Starcraft II \citep{Vinyals2019-sn} and Dota 2 \citep{OpenAI2019-ij}.

\subsection{World Models}\label{worldmodels}
Whereas model-free methods can directly learn from high-dimensional pixel images, MBRL faces the difficulty that a transition model is hard to train from such high-dimensional data. MBRL was shown to be efficient and successful for small sets of low-dimensional data, such as PILCO \citep{Deisenroth2015-jc}, where the authors used Gaussian processes (GP) to learn a model of the environment from which trajectories could be sampled. However, methods such as GP don't scale to high-dimensional observations. One work showed that by first using an autoencoder (AE) \citep{ballard1987modular} to learn a lower-dimensional representation of image data, a controller could be trained to balance a pendulum using the feature vectors provided by the bottle-neck hidden layer of the AE \citep{Wahlstrom2014-ym}. This usage of compressed latent spaces allowed much more sample efficient RL methods to be developed \citep{Finn2015-gt,Watter2015-zk}.

More recently, World Models \citep{Ha2018-jd} were introduced that allowed MBRL agents to learn policies for visual environments without having to interact with the environment. The authors first trained a Variational AE (VAE) \citep{rezende2014stochastic,kingma2014autoencoding} to learn a compressed representation of randomly gathered image data. Opposed to learning a function to compress the inputs, as is done with vanilla AEs, VAEs learn the parameters of a probability distribution representing the data. Using the VAE produced latent space, an RNN was trained that takes as input a latent state $z_{t}$, an action $a_t$, and a hidden state $h_t$ to predict a probability distribution of the next latent state $z_{t+1}$. The RNN is combined with a Mixture Density Network (MDN), such that a temperature parameter can be introduced that allows controlling the model's uncertainty when sampling from the distribution. After training, the MDN-RNN model allowed 'imagined' trajectories to be sampled that represent the actual environment. Additionally, the authors could assume access to the underlying reward functions of the used environments. A linear controller policy could therefore be learned without interacting with the actual environment, after which the policy was able to be transferred to the actual environment. The authors of SimPLe extended this approach, where they showed that a World Model can be used as a simulator to train model-free policies, resulting in impressive sample efficiency \citep{Kaiser2019-ts}.

Rather than learning a World Model and policy in separate stages, PlaNet \citep{Hafner2018-pq} trains these processes jointly and performs latent online planning for visual continuous control tasks. Moreover, instead of assuming the reward function of the environment, PlaNet learns a model of the reward as a function of the latent states. The latter allows planning to be performed completely in imagination. For planning, the authors combine model-predictive control with the cross entropy method. The latter is a population-based optimizer that iteratively evaluates a set of candidate action sequences under the model that are sampled from a belief distribution, which is repeatedly re-fit with the top $K$ rewarding action sequences. As transition model architecture, the authors introduce the Recurrent State Space Model (RSSM). This model combines a deterministic and stochastic model into one architecture. Purely deterministic models, such as a vanilla RNN, cannot capture multiple futures, and tend to exploit model inaccuracies. Purely stochastic models have difficulties remembering information over multiple time steps, given the randomness in transitions. The authors split the latent state in stochastic and deterministic parts, allowing robust prediction of multiple futures. This work provided the foundation for the state-of-the-art MBRL continuous control algorithm Dreamer \citep{Hafner2019-ab}, as well as its variant that is adapted for discrete problems, DreamerV2 \citep{Hafner2020-ry}. Dreamer will be the main focus of this work, and a detailed explanation of the core components and processes can  be found in Appendix \ref{dreamer}.

\newtheorem{definition}{Definition}
\subsection{Transfer Learning} \label{transfer}
This section introduces the concept of TL in the RL domain. We define TL and detail on how it is often used for RL algorithms (Section \ref{tlrl}). Additionally, we cover prior research that have applied TL to MBRL, with a main focus of approaches using World Models (Section \ref{wmtl}).

\subsubsection{Transfer learning in Context of Reinforcement Learning} \label{tlrl}
The idea of TL is to re-use information from a previous (source) task to enhance the learning of a new (target) task. The most frequent type of TL is parameter-based TL, meaning we transfer the parameters of a neural network. This is generally done by either freezing the pre-trained network (not updating its parameters), or retraining the pre-trained parameters, referred to as fine-tuning \citep{Yosinski2014-ze, DBLP:journals/corr/abs-2102-05207, sabatelli2018deep}. Moreover, the output layer of neural networks are generally randomly initialized in parameter-based TL, as it is already fit to produce outputs for a different task.

Formally, we can define TL in context of RL as follows (adapted from \cite{Pan2010-hg,Zhuang2019-rk,Zhu2020-jj}):
\begin{definition}
\normalfont (Transfer learning for reinforcement learning) Given a source task $T_S$ taking place in source MDP $\mathcal{M}_S$, as well as a target task $T_T$ taking place in target MDP $\mathcal{M}_T$, \textit{transfer learning for reinforcement learning} aims to enhance learning of the optimal policy $\pi^*$ in $\mathcal{M}_T$ by using knowledge of $\mathcal{M}_S$ and $T_S$, where $\mathcal{M}_S \neq \mathcal{M}_T$ or $T_S \neq T_T$. 
\end{definition}
That is, the objective function is the expected cumulative reward obtained by the optimal policy $\pi^*$, and the domains are limited to being MDPs. For simplicity we restrict this definition to single-source TL, but the number of source domains and tasks can be greater than one. When applying TL to RL, we need to consider what differences exist between the source and target MDPs. The main (partial) differences between domains can be in the state space $S$, the action space $A$, the reward function $R(s,a,s')$, and the transition dynamics $P(s,a,s')$. Additionally, two domains may differ in the initial states the agent starts in, or the number of steps an agent is allowed to move. Depending on the differences between the domains, we can consider what components to transfer from source to target domain. For instance, depending on the differences of the action spaces, state spaces, and reward functions, we can transfer policy and value functions \citep{Carroll2002-bh, Fernandez2006-tf}, as well as reward functions \citep{Schaal2004-ud}. One can also transfer experience samples that were collected during the training process of a set of source tasks, and use these to improve the learning of a given target task \citep{sampletransferlazaric,tirinzoni2018importance}.

TL is particularly popular in DRL, as, for example, pre-trained feature extractions of CNNs can often be reused to process high-dimensional image inputs of similar domains, which speeds up the learning process \citep{Zhu2016-cg}. Distillation techniques are a type of TL approach where a new network learns to predict the mappings of inputs to outputs of a pre-trained network \citep{Hinton2015-cw}. DRL researches have used this method to transfer policies from large DQNs, or multiple DQNs trained on different tasks, to a single and smaller DQN \citep{Parisotto2015-af, Rusu2015-bu}. Combinations of TL and multi-task learning can result in improvements of sample efficiency and robustness of current DRL algorithms \citep{Teh2017-gj}. The usage of neural networks also allows the transfer of feature representations, such as representations learned for value and policy functions \citep{Zhang2018-eo, Rusu2016-br}.

In order to evaluate the performance of TL approaches, several metrics have been proposed, such as the jumpstart (initial) performance, overall performance, and asymptotic (ultimate) performance \citep{Zhu2020-jj, Taylor2009-is}. By comparing these metrics of an agent trained with TL and without TL, we can evaluate the effects of a TL approach in different ways: mastery (ultimate performance), generalization (initial performance), and a combination of both (overall performance).

\subsubsection{Transfer Learning using World Models} \label{wmtl}
When applying TL to MBRL, we can also consider to transfer the dynamics model to the target domain if it contains similar dynamics to the source domain \citep{Eysenbach2020-to}. When using World Models, state observations are mapped to a compressed latent representation by means of reconstruction. In this latent space reward, dynamics, policy, and value models can be trained on relatively low-dimensional information. The authors in \cite{Zhu2020-jj} suggest that this can be beneficial for transferring knowledge across tasks that, for instance, share the environment dynamics, but have different reward functions. The authors of SEER show that for similar domains, the AE trained to map state observations to latent space can be frozen early-on in the training process, as well as transferred to new tasks and domains to save computation  \citep{chen2021improving}. PlaNet \citep{Hafner2018-pq} was used in a multi-task experiment, where the same PlaNet agent was trained on tasks of six different domains simultaneously. The agent infers which task it is doing based on the visual observations, and the authors show that a single agent becomes able to master these tasks using a single World Model. Due to the same physical laws that are present in physics engine used by the different domains, the transition model is able to transfer this knowledge among the different tasks. Similarly, the authors of \citep{Landolfi2019-mo} also perform a multi-task experiment, where the dynamics model of a MBRL agent is transferred to several novel tasks sequentially, and show that this results in significant gains of sample efficiency. In Plan2Explore \citep{Sekar2020-fl} the authors show that a single global World Model can be trained task-agnostically, after which a Dreamer \citep{Hafner2019-ab} agent can use the model to zero-shot or few-shot learn a new task. They show that this can result in state-of-the-art performance for zero-shot and few-shot learning, as well as competitive performance with a vanilla Dreamer agent.

\section{Fractional Transfer Learning} \label{fractransfmeth}
In this section we introduce the main contributions of this work, being transfer of simultaneous multi-task agents, and fractional transfer learning (FTL). The idea here is that we train a single agent on multiple tasks simultaneously (Section \ref{multi}), after which we transfer its parameters to a novel task. Moreover, unlike common TL approaches, we only transfer fractions of the parameters for certain components (Section \ref{fractransfsub}). Finally, we detail for each component of Dreamer whether it should receive no transfer, fractional transfer, or full transfer (Section \ref{appdreamftl}).

\subsection{Simultaneous Multi-Task Learning} \label{multi}
As was shown in PlaNet \citep{Hafner2018-pq}, a single agent using the RSSM World Model is able to learn multiple tasks from different domains simultaneously if the dynamics across the domains are similar, and visual observations are used to be able to infer which domain is being faced. In this work we investigate whether these simultaneous multi-task agents can serve as a new type of multi-source TL for RL.  In order to train agents on multiple tasks from different domains simultaneously, some adjustments need to be made to Dreamer, as was done in the PlaNet paper. First, when dealing with multiple tasks simultaneously, the action dimension $|A|$ of each domain may be different. Components like the action model in Dreamer require $|A|$ to be fixed in order to provide compatible outputs for each domain. To overcome this issue, we first compute which task has the largest action space dimensions, which is then used as the target dimension for padding the action spaces of all other tasks with unused elements:
\begin{equation}
    |A|_D = \text{max}(|A|_\textbf{D})
\end{equation}
Where $D \in \textbf{D}$ is domain $D$ from the set of domains $\textbf{D}$ that the agent will be trained in. This means the action space for each domain will be equivalent, but for tasks with originally smaller action spaces there are elements that have no effect on the environment and can therefore be ignored by the agent. Next, in order to acquire a balanced amount of gathered experience in the replay dataset, we apply the following modifications. Instead of randomly collecting 5 episodes for initialization, we randomly collect 1 episode of each task. Additionally, instead of collecting 1 episode of a single task after 100 training steps, we collect 1 episode for each of the training tasks.

\subsection{Fractional Transfer} \label{fractransfsub}
As discussed in Section \ref{transfer}, when transferring representations of neural networks the common approach is to re-use all weights of each layer. The final layer often needs to be randomly initialized for learning new tasks, as it otherwise needs to unlearn the previously fit weights. The other layers are either frozen, or retrained together with the final layer. However, the drawback of initializing a layer randomly is that all previously obtained knowledge contained in that layer is lost. Therefore, we present a novel alternative TL approach: FTL. The idea is to transfer a fraction of the weights of a source layer to a target layer. This can be done adding a fraction of the source layer's weights to the randomly initialized weights of a target layer. Rather than discarding all knowledge by random initialization, this approach allows the transfer of portions of potentially useful knowledge. Additionally, we can avoid overfitting scenarios with this approach as we don't necessarily fully transfer the parameters.

Formally, we can simply define FTL as:
\begin{equation}
    W_T = W_T + \omega W_S
\end{equation}
Where $W_T$ is the randomly initialized weights of a target layer, $W_S$ the trained weights of a source layer, and $\omega$ the fraction parameter that quantifies what proportion of knowledge is transferred, where $\omega \in [0,1]$. That is, for $\omega=0$ we have no transfer (i.e. random initialization), and for $\omega=1$ we have a full transfer of the weights. 

\begin{figure}[H]
    \centering
    \includegraphics[width=\textwidth]{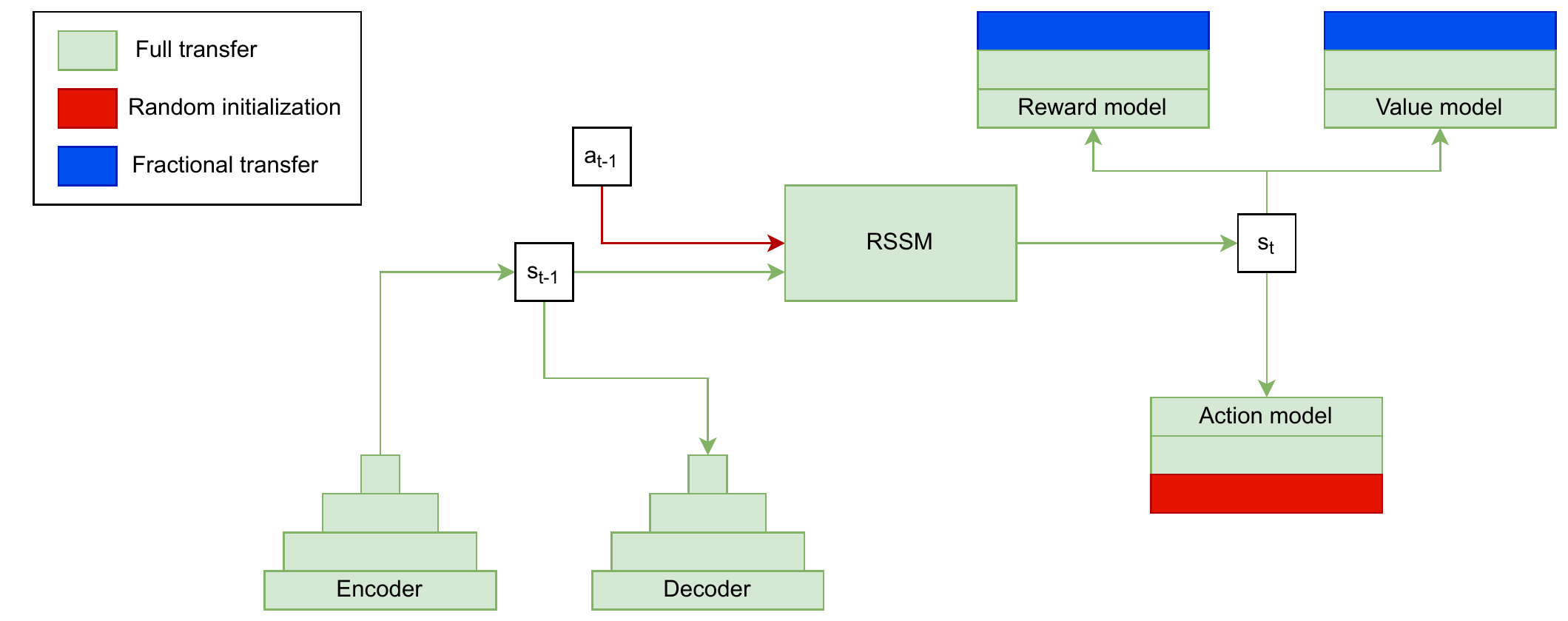}
    \caption{A schematic diagram of Dreamer's components and which type of TL is applied to each of the components. Green represents full transfer, red represents random initialization, blue represents FTL, arrows represent relevant input/output weights, and rectangles represent layers or models. For instance, the input weights of the RSSM model for an action $a_{t-1}$ are randomly initialized.}
    \label{fractransfdiag}
\end{figure}

\subsection{Application to Dreamer}\label{appdreamftl}
The experimental goal of this study is to transfer knowledge from agents that were simultaneously trained on multiple tasks to novel tasks. We are interested in observing whether this provides for a positive transfer, and additionally whether FTL provides for even better performances. However, as Dreamer consists of several different components, we need to identify what type of TL is most suitable for transferring each component's learned representation. In Figure \ref{fractransfdiag} a schematic illustration of all components of Dreamer and what type of transfer is applied to them in this work can be observed, as will be discussed below. The reward, value, and action models are each implemented as three fully-connected layers, the encoder and decoder are four-layer CNNs, and the dynamics model is the RSSM as described in Section \ref{worldmodels}. All of the following described parameters that are transferred are kept trainable afterwards. 

First, we identify the components of Dreamer that require random initialization, which are the components that involve processing actions. This is due to the fact that when transferring to a new domain, the action dimensions are likely incompatible with the action dimensions from the multi-source agents. Moreover, even if the action dimensions are equivalent, when the source and target domains are different, the learnt behaviours won't apply to the target domain and have to be unlearned when fully transferred. Therefore, we chose to randomly initialize (i.e. $\omega=0$) the last layer of the action model, and the weights that connect the input actions to the forward dynamics model. 

Next, we identify the components that are suitable for FTL, being the value model and reward model. For these neural networks we are not dealing with dimension mismatches, meaning random initialization is not a necessity. Additionally, full transfers are not applicable, as reward and value functions are generally different between the source and target domains, which would mean the weights of the source task would have to be unlearned. However, though these functions may not be equivalent across domains, the learned weights may still contain useful information that can be transferred with FTL. For these reasons we perform FTL for the last layer of the reward model and the last layer of the value model.

For the action model, value model, and reward model we only perform FTL and random initialization for the last layer. For the previous layers of these models we apply full transfer, as is generally done in TL for supervised learning, where only the last layer is randomly initialized. As the previous layers solely learn feature extractions of the latent state inputs, they can be re-used across domains. Preliminary empirical results showed that the best performances were obtained when only treating the last layer differently.

Finally, we identify which components of Dreamer are suitable for full transfers, which are the forward dynamics model and the VAE. The forward dynamics model can likely be fully transferred across domains if, for instance, the physical laws across the domains are equivalent, meaning this type of knowledge can be re-used \citep{Taylor2009-is}. The domains that will be used in this study share the same physics engine (Section \ref{exp}), meaning we decided to completely transfer the dynamics model parameters. We can also fully transfer the parameters of the representation model, as the generality of convolutional features allow reconstruction of observations for visually similar domains, and can therefore quickly adapt to novel domains \citep{chen2021improving}.

\begin{figure}[H]
    \centering
    \includegraphics[scale=0.325]{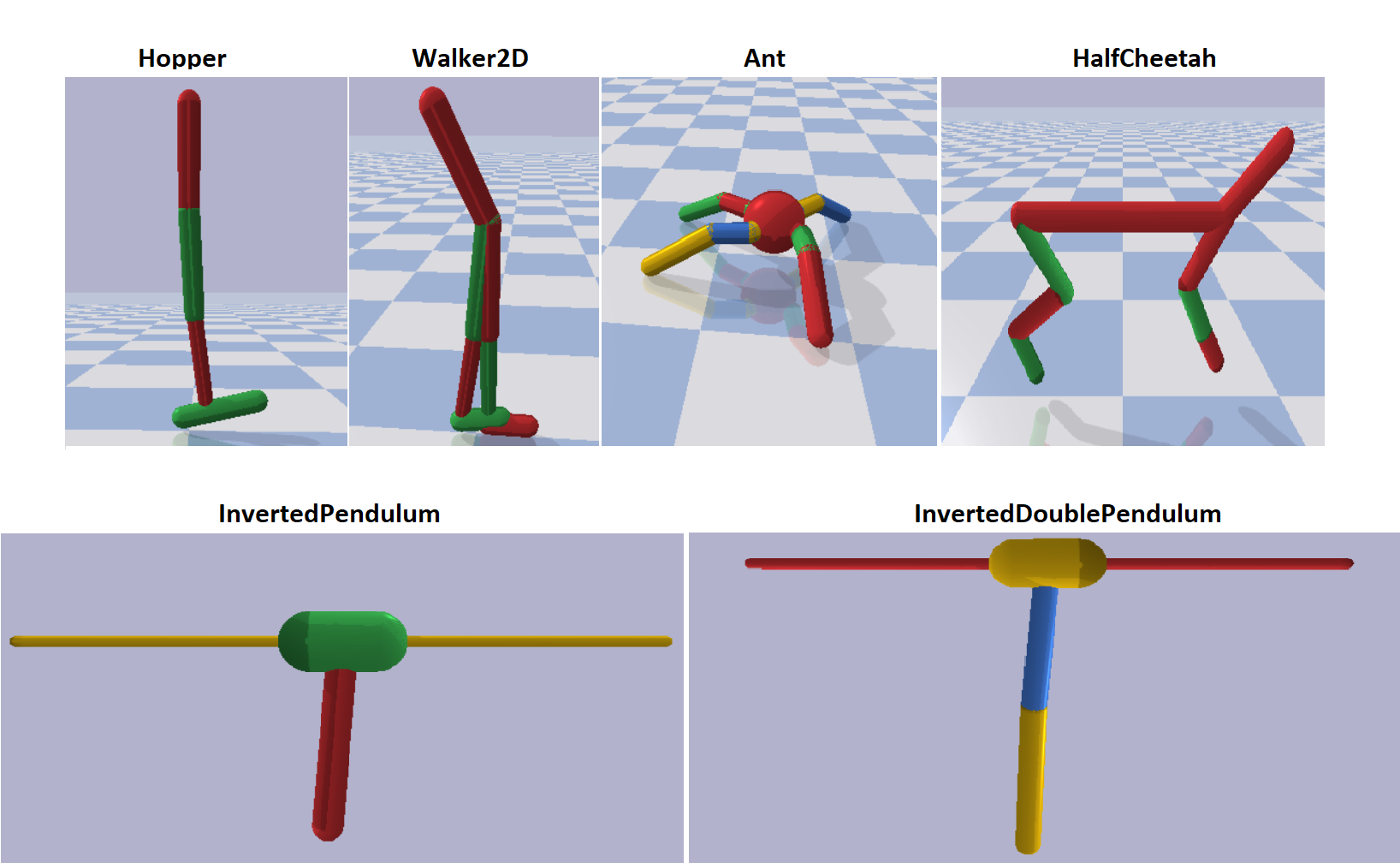}
    \caption{The four locomotion tasks (top) and two pendulum swingup tasks (bottom) provided by PyBullet that were used in the experiments of this research.}
    \label{fig:tasks}
\end{figure}

\section{Experiments} \label{exp}
In order to perform experiments for the proposed methodologies, we chose to use several continuous control tasks from the environments that are provided by the PyBullet physics engine. These environments are very similar to the continuous control environments provided by the OpenAI Gym \citep{Brockman2016-ri}, yet are open-source and harder to solve \citep{Raffin2020-ly}. For the experiments we chose a set of 4 locomotion tasks and 2 pendula swing-up tasks (Figure \ref{fig:tasks}). For each of the environments and experiments we provide image observations from the simulation to the algorithms, and return task-compatible action vectors to perform actions.

The set of source tasks and the set of target tasks used in this experiment consist of the Hopper, Ant, Walker2D, HalfCheetah, InvertedPendulum, and InvertedDoublePendulum\footnote{PyBullet does not provide a swing-up task for the InvertedDoublePendulum environment. Therefore, we customized this ourselves by modifying the reward function. For details see \cite{Sasso2021}.} environments. First, a single agent is trained on multiple source tasks as described in Section \ref{multi}. For each target task we train simultaneous multi-task agents of 2, 3, and 4 source tasks. For a complete overview of which source and target agents combinations were used, see Appendix \ref{expdetails}.

Each of the source agents is trained for 2e6 environment steps for a single run, where 1 episode of each task is collected when gathering data. For each source agent we transfer the weights of all components to the corresponding target agent, except for the weights of the last layer of the reward, value, actor networks, as well as the action-to-model weights. For this paper, we added weight fractions of $\omega = 0.2$ from the source to the randomly initialized weights of the last layer of the reward and value networks. However, note that we do not claim this to be the optimal fraction to be used in all cases. As can be seen in Appendix \ref{morefracresults}, for different testing environments and source-target combinations the transfer fraction resulting in the best performance varies. $\omega$ can therefore be considered as a tunable hyperparameter, and based on our previous study we can recommend $\omega \in [0.1,0.4]$ \citep{Sasso2021}. For this paper we chose $\omega = 0.2$, as across the testing environments used in this study we observed the largest performance gains for $\omega \in [0.1,0.3]$. 

Each source-target combination trains for 1e6 environment steps for three random seeds. We evaluate the target agents by obtaining an episode return every 1e4 environment steps of the training process. We compare the overall average episode return, as well as the average episode return of the final 1e5 environment steps, to a baseline Dreamer agent that was trained from scratch for three random seeds on each of the target tasks.

\section{Results}\label{results}
\begin{figure}[H]
    \centering
    \includegraphics[, width=\linewidth]{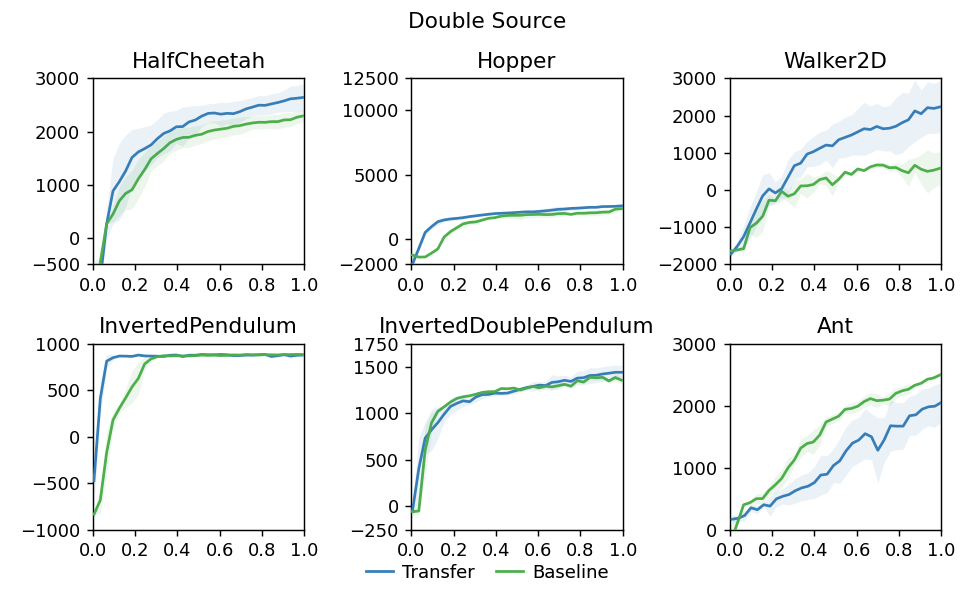}
    \caption{Episode return across 3 random seeds of having simultaneously trained on two source tasks, after which fractional transfer was done with a fraction $\omega = 0.2$ transferred to the HalfCheetah, Hopper, Walker2D, InvertedPendulum, InvertedDoublePendulum, and Ant tasks.}
    \label{fig:2task}
\end{figure}

\begin{figure}[H]
    \centering
    \includegraphics[, width=\linewidth]{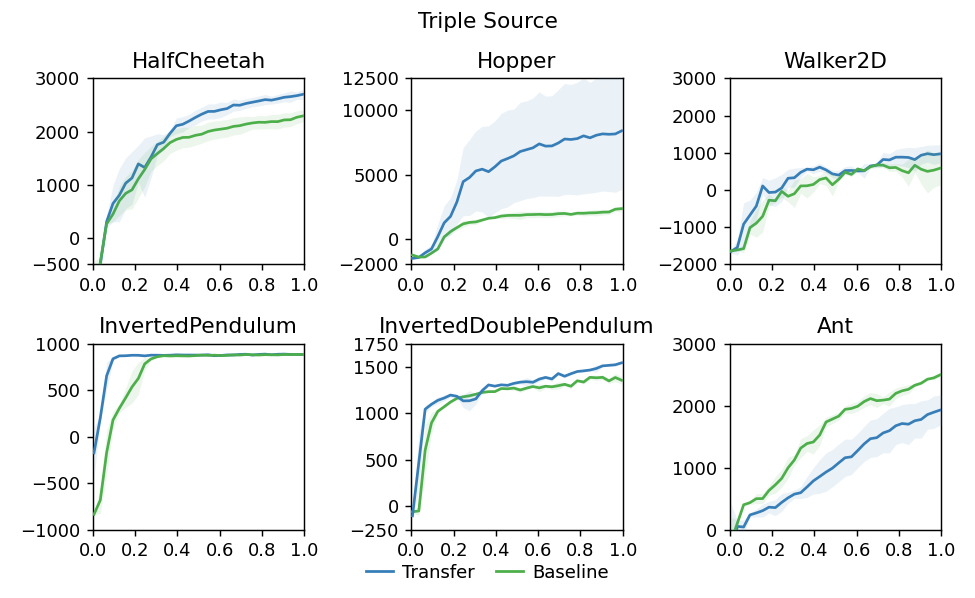}
    \caption{Episode return across 3 random seeds of having simultaneously trained on three source tasks, after which fractional transfer was done with a fraction $\omega = 0.2$ transferred to the HalfCheetah, Hopper, Walker2D, InvertedPendulum, InvertedDoublePendulum, and Ant tasks.}
    \label{fig:3task}
\end{figure}

\begin{figure}[H]
    \centering
    \includegraphics[, width=\linewidth]{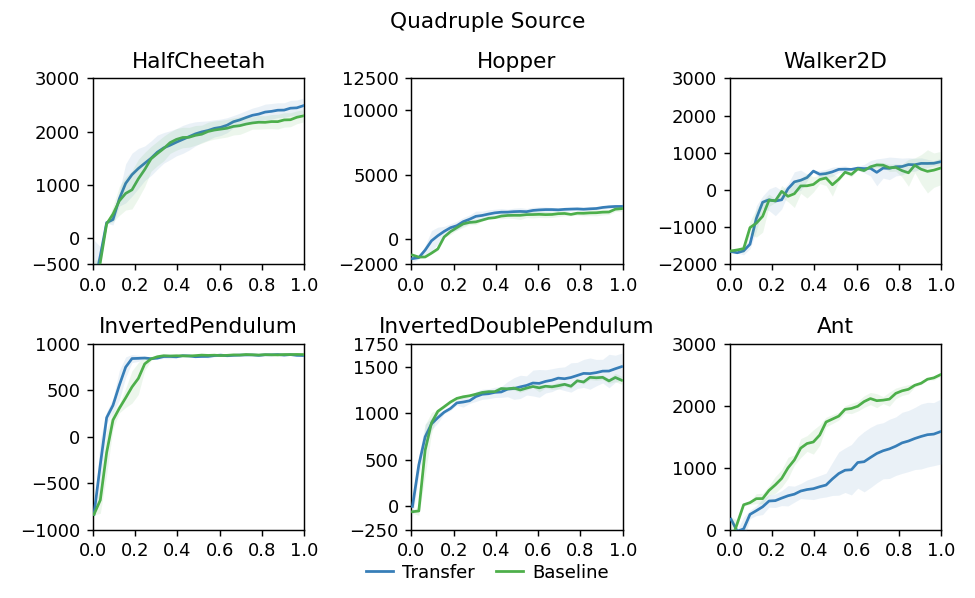}
    \caption{Episode return across 3 random seeds of having simultaneously trained on four source tasks, after which fractional transfer was done with a fraction $\omega = 0.2$ transferred to the HalfCheetah, Hopper, Walker2D, InvertedPendulum, InvertedDoublePendulum, and Ant tasks.}
    \label{fig:4task}
\end{figure}

In this section we present the results for FTL. The results for 2, 3, and 4 source tasks using $\omega = 0.2$ for each of the testing environments can be seen in Figure \ref{fig:2task}, Figure \ref{fig:3task}, and Figure \ref{fig:4task} respectively. In each figure the overall performance of the transfer learning agent (blue) can be seen compared to the overall performance of a baseline agent (green), where for both the standard deviation of three runs is represented by the shaded area. Table \ref{table:overalltable} and Table \ref{table:finaltable} contain the corresponding numerical results for the overall performance, and the performance of the final 1e5 environment steps respectively.

The performances for the same source agents with $\omega = 0.0$ (random initialization) can be found in Appendix \ref{randominit}. A set of additional results for $\omega = [0.0,0.1,0.2,0.3,0.4,0.5]$ can be found in Appendix \ref{morefracresults}. For a complete overview of all the experimental results for each of the fractions applied to all testing environments we refer the reader to the thesis this study originates from \citep{Sasso2021}.

\begin{table}[H]
    \centering
    \begin{tabular}{c|ccc|c}
        & 2 Tasks & 3 Tasks & 4 Tasks & Baseline \\
        \hline
        HalfCheetah  & \textbf{1982 $\pm$ 838} & 1967 $\pm$ 862 & 1773 $\pm$ 748& 1681 $\pm$ 726\\
        Hopper & 1911 $\pm$ 712 & \textbf{5538 $\pm$ 4720} & 1702 $\pm$ 1078 &  1340 $\pm$ 1112\\
        Walker2D & \textbf{1009 $\pm$ 1254} & 393 $\pm$ 813 & 200 $\pm$ 807 & 116 $\pm$ 885\\
        InvertedPendulum & 731 $\pm$ 332 & \textbf{740 $\pm$ 333} & 731 $\pm$ 302 & 723 $\pm$ 364\\
        InvertedDoublePendulum & 1209 $\pm$ 280 & \textbf{1299 $\pm$ 254} & 1235 $\pm$ 284 & 1194 $\pm$ 306\\
        Ant & 1124 $\pm$ 722 & 1052 $\pm$ 687 & 898 $\pm$ 616 & \textbf{1589 $\pm$ 771} \\
    \end{tabular}
    \caption{Average episode return of fractional transfer learning for $\omega = 0.2$, using 2, 3, and 4 source tasks transferred to the HalfCheetah, Hopper, Walker2D, InvertedPendulum, InvertedDoublePendulum, and Ant tasks.}
    \label{table:overalltable}
    
\end{table}
\begin{table}[H]
    \centering
    \begin{tabular}{c|ccc|c}
        & 2 Tasks & 3 Tasks & 4 Tasks & Baseline \\
        \hline
        HalfCheetah & \textbf{2820 $\pm$ 297} & 2615 $\pm$ 132 & 2276 $\pm$ 240 & 2264 $\pm$ 160\\
        Hopper &2535 $\pm$ 712 & \textbf{8274 $\pm$ 4649 }& 2507 $\pm$ 240& 2241 $\pm$ 502 \\
        Walker2D &\textbf{ 2214 $\pm$ 1254} & 963 $\pm$ 314 & 733 $\pm$ 183 & 547 $\pm$ 710\\
        InvertedPendulum &874 $\pm$ 121 &\textbf{ 884 $\pm$ 20} & 878 $\pm$ 34 & 883 $\pm$ 17\\
        InvertedDoublePendulum & 1438 $\pm$ 116 & \textbf{1531 $\pm$ 93} & 1481 $\pm$ 188 & 1366 $\pm$ 179 \\
        Ant &  2021 $\pm$ 722 & 1899 $\pm$ 292 & 1556 $\pm$ 520 & \textbf{2463 $\pm$ 208} \\
    \end{tabular}
    \caption{Final averaged episode return of the last 1e5 environment steps fractional transfer learning for $\omega = 0.2$, using 2, 3, and 4 source tasks transferred to the HalfCheetah, Hopper, Walker2D, InvertedPendulum, InvertedDoublePendulum, and Ant tasks.}
    \label{table:finaltable}
\end{table}

\raggedbottom

\section{Discussion}
We found that using the state-of-the-art deep MBRL algorithm for continuous control tasks, simultaneously training on multiple tasks results in a positive transfer in most cases, both in terms of overall performance and ultimate performance. However, depending on the number of relevant source tasks, this approach may also result in a neutral or negative transfer. For instance, for the HalfCheetah, Hopper, and Walker2D locomotion testing environments we generally observe a decrease in both overall and ultimate performance as we add more tasks that provide irrelevant information to the target domain, such as the pendula environments (Table \ref{table:0overalltable} and Table \ref{table:0finaltable}). However, the quantity of source tasks does not influence the performance, as can be seen in the results of the pendula testing environments. Here, the third and fourth added tasks are both irrelevant, yet 3 source tasks does not necessarily outperform 4 source tasks. In these experiments the pendula environments are the only relevant sources for each other, and the results therefore also suggest that as long as there is at least one relevant source task in the multi-source collection, a positive transfer is likely to occur. Lastly, for the Ant testing environment we observe significant negative transfers for any set of source tasks, which is likely due to the movement dynamics and reward function being too different from the other domains, meaning all source tasks provide irrelevant (and therefore interfering) information. 

We also showed that we can boost the performance of the aforementioned TL technique by transferring fractions of parameters of neural network layers, rather than transferring all parameters or random initialization. This can be done by adding a fraction of the source parameters to randomly initialized parameters. We identified that this type of TL is beneficial for DRL components such as the reward model and value model, as these components suffer from overfitting when fully transferred, and do not require to be re-initialized as they are unrelated to actions. Compared to randomly initializing last layers or learning from scratch, transferring a fraction of $\omega = 0.2$ results in significantly better overall and ultimate performance for almost all testing environments, except for the Ant environment as argued for earlier (Table \ref{table:overalltable} and Table \ref{table:finaltable}).

\section{Conclusion \& Future Work}
In this study we introduced a novel TL method for RL algorithms: FTL. Using Dreamer, we trained several single agents on multiple tasks simultaneously, after which an agent of the like was tasked with learning a novel task. We identified which components of Dreamer could be fully transferred, and which components require random initialization of their output layer to be able to adapt to the new task. The initial results suggested that the used multi-source transfer learning technique results in positive transfers when using World Model-based architectures. We then showed that, instead of randomly initializing the output layer, transferring fractions of parameters to the output layer of the reward and value models results in significant performance improvements. 

We applied FTL in a relatively complex setting, as we used a sophisticated deep MBRL algorithm to simultaneously train a single agent on multiple visual tasks. This suggests that FTL should also be applicable to other algorithms and single-source TL settings, which is therefore a promising direction of future research. Additionally, in this research we fully transferred the parameters of all layers of the dynamics model of Dreamer. However, we saw that this may result in a much worse performance when the dynamics are too different. It would therefore be interesting to apply FTL to the dynamics model as well, such that just a small portion of dynamics knowledge is transferred, likely preventing the interference observed in this study. Lastly, it could be interesting to investigate whether FTL is also beneficial for applications in the supervised learning domain.
\raggedbottom

\newpage
\bibliography{biblio}
\bibliographystyle{iclr2021_conference}
\newpage
\appendix

\section{Dreamer} \label{dreamer}
In this section we detail on the core components and processes of Dreamer. For complete details we refer the reader to \cite{Hafner2019-ab}. Dreamer is a state-of-the-art deep MBRL algorithm for continuous control tasks. Dreamer learns using the following processes:
\begin{itemize}
    \item A latent dynamics model is learned from a dataset of past experience to predict future rewards and transitions of actions and observations.
    \item An action and value model are learned from imagined latent trajectories.
    \item The action model is used in the environment to collect experiences that are added to the growing dataset.
\end{itemize}
The overall process of Dreamer is that it repeatedly collects experience data, learns latent dynamics using a World Model, and then uses this model to learn value and policy functions by imagining trajectories in latent space. The latent dynamics model is the RSSM architecture from PlaNet \citep{Hafner2018-pq}. The World Model consists of three components:
\begin{itemize}
    \item Representation model $p_\theta(s_t|s_{t-1},a_{t-1},o_t)$, that maps state observations and actions to vector-valued latent states.
    \item Transition model $q_\theta(s_t|s_{t-1},a_{t-1})$, that predicts the next latent state given a latent state and action, without having to observe or imagine the corresponding images.
    \item Reward model $q_\theta(r_t|s_{t})$, that predicts the reward of a latent state.
\end{itemize}
Here $p$ denotes distributions generating real environment samples, $q$ denotes Gaussian distributions generating imagined samples, and $\theta$ denotes the neural network parameters of the models. The representation model is a variational auto-encoder (VAE) that is trained by reconstruction. The transition model learns by comparing its prediction of a latent state-action transition to the corresponding encoded next true state. The reward model learns to predict rewards of latent states by using the true reward of the environment state as a target. The imagined trajectories start at true model states $s_t$ sampled from approximate state posteriors yielded by the representation model $s_t \sim p_\theta(s_t|s_{t-1},a_{t-1},o_t)$, which are based on past observations $o_t$ of the dataset of experience. The imagined trajectory then follows predictions from the transition model, reward model, and a policy, by sampling $s_\tau \sim q_\theta(s_\tau|s_{\tau-1},a_{\tau-1})$, taking the mode $r_\tau \sim q_\theta(r_\tau|s_{\tau})$, and sampling $a_\tau \sim q_\phi(a_\tau|s_\tau)$ respectively, where $\tau$ denotes a time step in imagination.

In order to learn policies, Dreamer uses an actor-critic approach, which is a popular reinforcement learning algorithm for continuous spaces. An actor-critic consists of an action model and a value model, where the action model implements the policy, and the value model estimates the expected reward that the action model achieves from a given state:
\begin{itemize}
    \item Actor model: $q_\phi(a_\tau|s_\tau)$
    \item Value model: $v_\psi(s_t) \approx E_{q(\cdot|s_\tau)}(\sum_{\tau=t}^{t+H}\gamma^{\tau-t}r_\tau)$
\end{itemize}
Here $H(=15)$ denotes the finite horizon of an imagined trajectory, $\phi$ and $\psi$ the neural network parameters for the action and value model respectively, and $\gamma(=0.99)$ is the discount factor. In order to learn these models, estimates of state values for the imagined trajectories $\left\{s_{\tau}, a_{\tau}, r_{\tau}\right\}_{\tau=t}^{t+H}$ are used. Dreamer acquires these by using an exponentially-weighted average of the estimates $V_\lambda$:
\begin{equation} \label{valuest}
    \begin{aligned}
        \mathrm{V}_{\mathrm{N}}^{k}\left(s_{\tau}\right) &\doteq \mathrm{E}_{q_{\theta}, q_{\phi}}\left(\sum_{n=\tau}^{h-1} \gamma^{n-\tau} r_{n}+\gamma^{h-\tau} v_{\psi}\left(s_{h}\right)\right) \quad  \text{with} \quad h=\min (\tau+k, t+H)\\
        \mathrm{V}_{\lambda}\left(s_{\tau}\right) &\doteq(1-\lambda) \sum_{n=1}^{H-1} \lambda^{n-1} \mathrm{~V}_{\mathrm{N}}^{n}\left(s_{\tau}\right)+\lambda^{H-1} \mathrm{~V}_{\mathrm{N}}^{H}\left(s_{\tau}\right)
    \end{aligned}
\end{equation}
That is, $V^k_N$ estimates rewards beyond $k$ imagination steps with the current value model. This is used in $V_\lambda$, where $\lambda(=0.95)$ is a parameter that controls the exponentially-weighted average of the estimates for different $k$. This value estimate is then used in the objective functions of the action and value models:
\begin{equation}
    \begin{aligned}
        &\max _{\phi} \mathrm{E}_{q_{\theta}, q_{\phi}}\left(\sum_{\tau=t}^{t+H} \mathrm{~V}_{\lambda}\left(s_{\tau}\right)\right)\\
        &\min _{\psi} \mathrm{E}_{q_{\theta}, q_{\phi}}\left(\sum_{\tau=t}^{t+H} \frac{1}{2} \| v_{\psi}\left(s_{\tau}\right)-\mathrm{V}_{\lambda}\left(s_{\tau}\right) \|^{2}\right)
    \end{aligned}
\end{equation}
That is, the objective function for the action model $q_\phi(a_\tau|s_\tau)$ is to predict actions that result in state trajectories with high value estimates. The objective function for the value model $v_\psi(s_\tau)$ is to regress these value estimates. As can be inferred from these definitions, the value model and reward model are crucial components in the behaviour learning process, as they determine the value estimation. The learning procedure of the action model is therefore completely dependent on the predictions of these models. Moreover, as the reward model and value model are dependent on imagined states of a fixed World Model that result from actions, they are also dependent on the action model. Each of the models mentioned are implemented as neural networks, and are optimized using stochastic backpropagation of multi-step returns. 

\newpage

\section{Experiment setup details}\label{expdetails}
The following shows what type of simultaneous multi-task agents (Section \ref{multi}) were trained for a given testing environment, as used in the experiments described in Section \ref{exp}.

The HalfCheetah multi-source agents are:
\begin{itemize}
    \item Hopper + Ant
    \item Hopper + Ant + Walker2D
    \item Hopper + Ant + Walker2D + InvertedPendulum
\end{itemize}
The Hopper multi-source agents are:
\begin{itemize}
    \item HalfCheetah + Walker2D
    \item HalfCheetah + Walker2D + Ant
    \item HalfCheetah + Walker2D + Ant + InvertedPendulum
\end{itemize}
The Walker2D multi-source agents are:
\begin{itemize}
    \item HalfCheetah + Hopper
    \item HalfCheetah + Hopper + Ant
    \item HalfCheetah + Hopper + Ant + InvertedPendulum
\end{itemize}
The InvertedPendulum multi-source agents are:
\begin{itemize}
    \item HalfCheetah + InvertedDoublePendulum
    \item HalfCheetah + InvertedDoublePendulum + Hopper
    \item HalfCheetah + InvertedDoublePendulum + Hopper + Ant
\end{itemize}
The InvertedDoublePendulum multi-source agents are:
\begin{itemize}
    \item Hopper + InvertedPendulum
    \item Hopper + InvertedPendulum + Walker2D
    \item Hopper + InvertedPendulum + Walker2D + Ant
\end{itemize}
The Ant multi-source agents are:
\begin{itemize}
    \item HalfCheetah + Walker2D
    \item HalfCheetah + Walker2D + Hopper
    \item HalfCheetah + Walker2D + Hopper + InvertedPendulum
\end{itemize}

\newpage
\section{Random initialization results} \label{randominit}
\begin{figure}[H]
    \centering
    \includegraphics[, width=\linewidth]{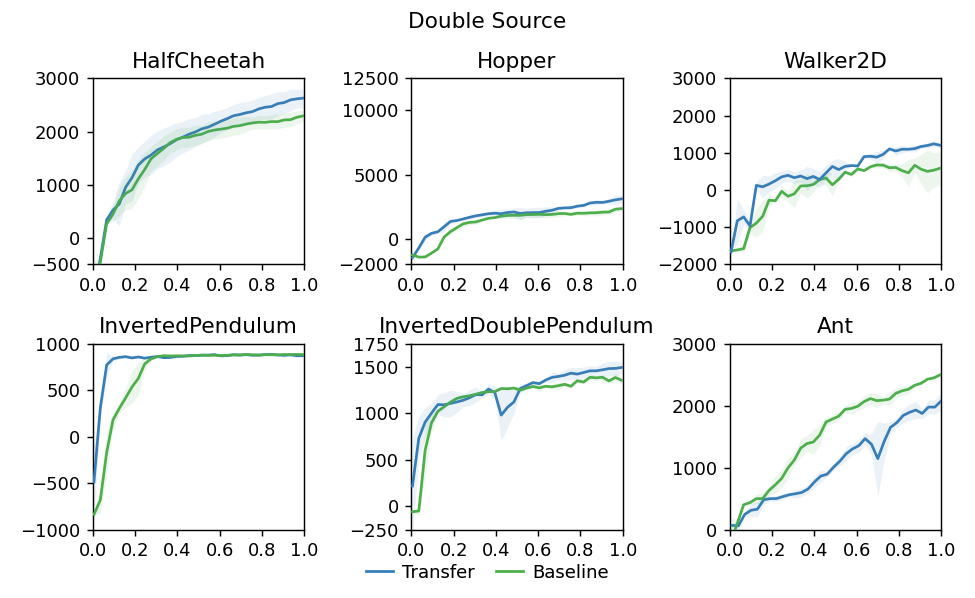}
    \caption{Episode return across 3 random seeds of having simultaneously trained on two source tasks, after which fractional transfer was done with a fraction $\omega = 0.0$ transferred to the HalfCheetah, Hopper, Walker2D, InvertedPendulum, InvertedDoublePendulum, and Ant tasks.}
    \label{fig:02task}
\end{figure}

\begin{figure}[H]
    \centering
    \includegraphics[, width=\linewidth]{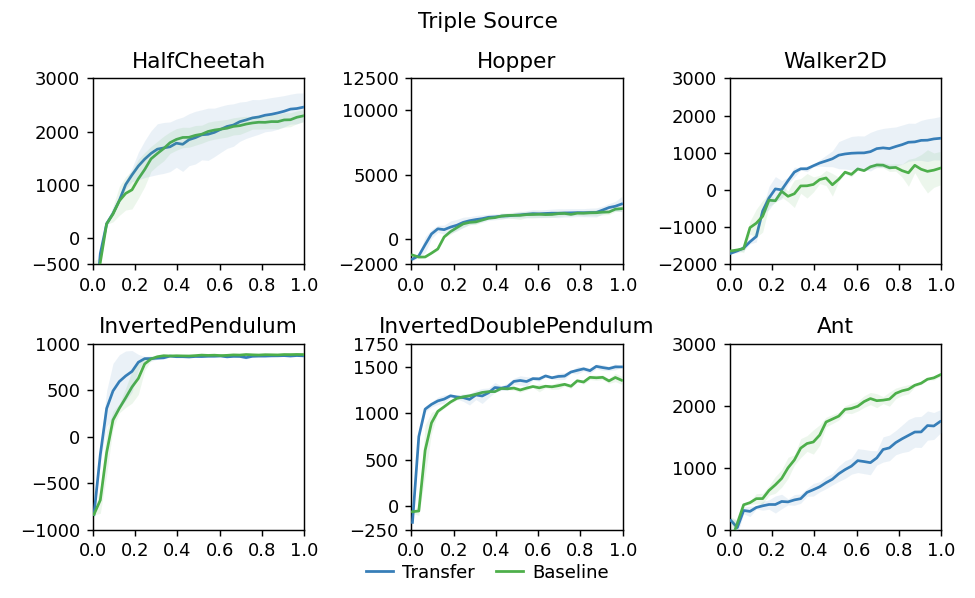}
    \caption{Episode return across 3 random seeds of having simultaneously trained on three source tasks, after which fractional transfer was done with a fraction $\omega = 0.0$ transferred to the HalfCheetah, Hopper, Walker2D, InvertedPendulum, InvertedDoublePendulum, and Ant tasks.}
    \label{fig:03task}
\end{figure}

\begin{figure}[H]
    \centering
    \includegraphics[, width=\linewidth]{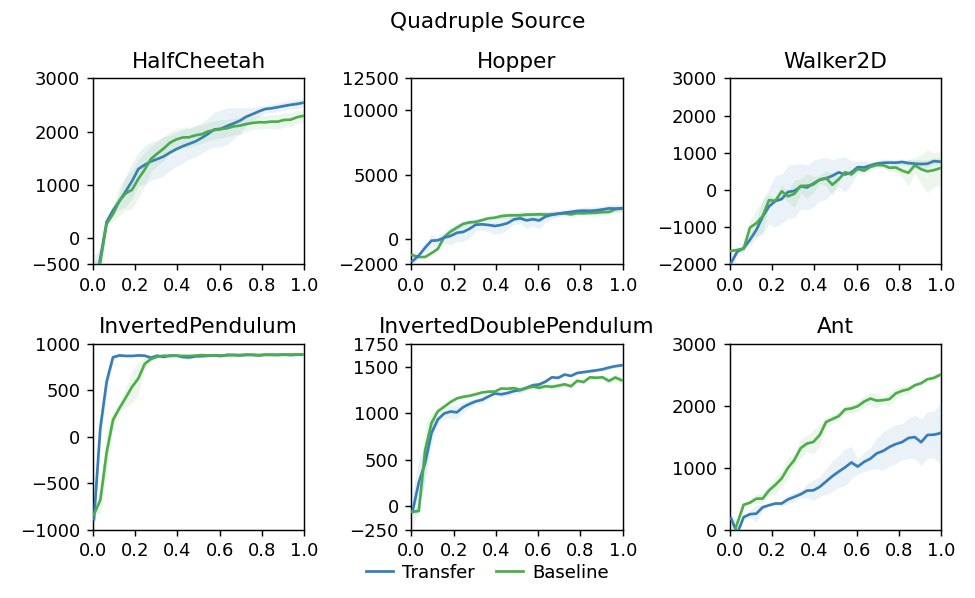}
    \caption{Episode return across 3 random seeds of having simultaneously trained on four source tasks, after which fractional transfer was done with a fraction $\omega = 0.0$ transferred to the HalfCheetah, Hopper, Walker2D, InvertedPendulum, InvertedDoublePendulum, and Ant tasks.}
    \label{fig:04task}
\end{figure}

Here the results corresponding to Section \ref{results} can be found for $\omega = 0.0$, i.e. random initialization of the output layer parameters of the value and reward models. The results for 2, 3, and 4 source tasks for each of the testing environments can be seen in Figure \ref{fig:02task}, Figure \ref{fig:03task}, and Figure \ref{fig:04task} respectively. Again, in each figure the overall performance of the transfer learning agent (blue) can be seen compared to the overall performance of a baseline agent (green), where for both the standard deviation of three runs is represented by the shaded area. Table \ref{table:0overalltable} and Table \ref{table:0finaltable} contain the corresponding numerical results for the overall performance, and the performance of the final 1e5 environment steps respectively.

\begin{table}[H]
    \centering
    \begin{tabular}{c|ccc|c}
        & 2 Tasks & 3 Tasks & 4 Tasks & Baseline \\
        \hline
        HalfCheetah  &\textbf{ 1841 $\pm$ 806} & 1752 $\pm$ 783  & 1742 $\pm$ 777 & 1681 $\pm$ 726\\
        Hopper & \textbf{1917 $\pm$ 960 }& 1585 $\pm$ 941 & 1263 $\pm$ 1113 &  1340 $\pm$ 1112\\
        InvertedPendulum & \textbf{ 847 $\pm$ 135} & 779 $\pm$ 259 & 838 $\pm$ 169 & 723 $\pm$ 364\\
        Walker2D & \textbf{546 $\pm$ 776} & 545 $\pm$ 1012 & 157 $\pm$ 891 & 116 $\pm$ 885\\
        InvertedDoublePendulum & 1255 $\pm$ 270 & \textbf{ 1303 $\pm$ 248} & 1208 $\pm$ 322 & 1194 $\pm$ 306\\
        Ant & 1070 $\pm$ 659 & 923 $\pm$ 559 & 897 $\pm$ 586 & \textbf{1589 $\pm$ 771} \\
    \end{tabular}
    \caption{Average episode return of fractional transfer learning for $\omega = 0.0$, using 2, 3, and 4 source tasks transferred to the HalfCheetah, Hopper, Walker2D, InvertedPendulum, InvertedDoublePendulum, and Ant tasks.}
    \label{table:0overalltable}
    
\end{table}
\begin{table}[H]
    \centering
    \begin{tabular}{c|ccc|c}
        & 2 Tasks & 3 Tasks & 4 Tasks & Baseline \\
        \hline
        HalfCheetah &\textbf{ 2614 $\pm$ 208} & 2435 $\pm$ 287 & 2520 $\pm$ 103& 2264 $\pm$ 160\\
        Hopper &\textbf{3006 $\pm$ 960} & 2560 $\pm$ 591 & 2357 $\pm$ 351& 2241 $\pm$ 502 \\
        InvertedPendulum & 875 $\pm$ 135 & 871 $\pm$ 38 &881 $\pm$ 21& \textbf{883 $\pm$ 17}\\
        Walker2D & 1212 $\pm$ 776 & \textbf{1368 $\pm$ 613} & 744 $\pm$ 256 & 547 $\pm$ 710\\
        InvertedDoublePendulum & 1489 $\pm$ 133 & 1494 $\pm$ 142 & \textbf{1506 $\pm$ 74} & 1366 $\pm$ 179 \\
        Ant &  1963 $\pm$ 659 & 1703 $\pm$ 256 & 1555 $\pm$ 515 & \textbf{2463 $\pm$ 208} \\
    \end{tabular}
    \caption{Final averaged episode return of the last 1e5 environment steps fractional transfer learning for $\omega = 0.0$, using 2, 3, and 4 source tasks transferred to the HalfCheetah, Hopper, Walker2D, InvertedPendulum, InvertedDoublePendulum, and Ant tasks.}
    \label{table:0finaltable}
\end{table}

\newpage

\section{Additional fraction results}\label{morefracresults}
In this appendix results can be found that illustrate the effect of different fractions on different testing environments and source-target combinations. We present the numerical results of transferring fractions of $\omega = [0.0,0.1,0.2,0.3,0.4,0.5]$ using 2, 3, and 4 source tasks. That is, the overall performance over 3 random seeds for the HalfCheetah, Hopper, and InvertedDoublePendulum testing environments can be found in Table \ref{table:fractransfcheetahtable}, Table \ref{hoppertable}, and Table \ref{pendulumtable} respectively.

For a complete overview of all the experimental results for each of the fractions applied to all testing environments we refer the reader to the thesis this study originates from \citep{Sasso2021}.

\begin{table}[H]
    \centering
    \begin{tabular}{c|ccc} 
        & 2 Tasks & 3 Tasks & 4 Tasks \\
        \hline
        0.0 & 1841 $\pm$ 806 & 1752 $\pm$ 783  & 1742 $\pm$ 777\\
        0.1 & \textbf{2028 $\pm$ 993} & 2074 $\pm$ 762 & 1500 $\pm$ 781\\
        0.2 & 1982 $\pm$ 838 & 1967 $\pm$ 862 & 1773 $\pm$ 748\\
        0.3 & 1899 $\pm$ 911 & \textbf{2094 $\pm$ 859} & 1544 $\pm$ 771\\
        0.4 & 2008 $\pm$ 943 & 2015 $\pm$ 873 & \textbf{2162 $\pm$ 789}\\
        0.5 & 1961 $\pm$ 944 & 1647 $\pm$ 896 & 1635 $\pm$ 809\\
        \hline
        Baseline & & 1681 $\pm$ 726 &
    \end{tabular}
    \caption{Average return for fraction transfer of 2, 3, and 4 source tasks (Hopper, Ant, Walker2D and InvertedPendulum) for the HalfCheetah task with fractions of 0.0 up to 0.5.}
    \label{table:fractransfcheetahtable}
\end{table}

\begin{table}[H]
    \centering
    \begin{tabular}{c|ccc}
        & 2 Tasks & 3 Tasks & 4 Tasks \\
        \hline
        0.0 & 1917 $\pm$ 960 & 1585 $\pm$ 941 & 1263 $\pm$ 1113\\
        0.1 & 2041 $\pm$ 887 & \textbf{6542 $\pm$ 4469} & 3300 $\pm$ 3342\\
        0.2 & 1911 $\pm$ 712 & 5538 $\pm$ 4720 & 1702 $\pm$ 1078\\
        0.3 & \textbf{2670 $\pm$ 1789} & 4925 $\pm$ 4695 & \textbf{3341 $\pm$ 4609}\\
        0.4 & 2076 $\pm$ 803 & 2437 $\pm$ 2731 & 1451 $\pm$ 991\\
        0.5 & 1975 $\pm$ 772 & 2014 $\pm$ 2328 & 3246 $\pm$ 3828\\
        \hline
        Baseline & &  1340 $\pm$ 1112 &
    \end{tabular}
    \caption{Average return for fraction transfer of 2, 3, and 4 source tasks (Cheetah, Walker2D, Ant and InvertedPendulum) for the Hopper task with fractions of 0.0 up to 0.5.}
    \label{hoppertable}
\end{table}

\begin{table}[H]
    \centering
    \begin{tabular}{c|ccc}
        & 2 Tasks & 3 Tasks & 4 Tasks \\
        \hline
        0.0 & 1255 $\pm$ 270 &  1303 $\pm$ 248& 1208 $\pm$ 322\\
        0.1 & 1185 $\pm$ 316 & 1283 $\pm$ 272 & 1251 $\pm$ 272 \\
        0.2 & 1209 $\pm$ 280 & 1299 $\pm$ 254 & 1235 $\pm$ 284 \\
        0.3 & 1279 $\pm$ 216 & 1325 $\pm$ 225 & 1239 $\pm$ 259\\
        0.4 & \textbf{1307 $\pm$ 227} & 1323 $\pm$ 243 & \textbf{1278 $\pm$ 287}\\
        0.5 & 1301 $\pm$ 216 & \textbf{1340 $\pm$ 230} & 1206 $\pm$ 278\\
        \hline
        Baseline & & 1194 $\pm$ 306 &
    \end{tabular}
    \caption{Average return for fraction transfer of 2, 3, and 4 source tasks (Hopper, InvertedPendulum, Walker2D and Ant) for the InvertedDoublePendulum task with fractions of 0.0 up to 0.5.}
    \label{pendulumtable}
\end{table}

\end{document}

%% file: math_commands.tex
\usepackage{amsmath,amsfonts,bm}

\def\eqref#1{equation~\ref{#1}}

\def\1{\bm{1}}

\DeclareMathAlphabet{\mathsfit}{\encodingdefault}{\sfdefault}{m}{sl}
\SetMathAlphabet{\mathsfit}{bold}{\encodingdefault}{\sfdefault}{bx}{n}

